\documentclass[runningheads]{llncs}
\usepackage{graphicx}
\usepackage[sort, numbers]{natbib}
\usepackage{amsfonts}
\usepackage{amsmath}
\usepackage{amssymb}
\usepackage{float}
\usepackage{subfig}
\usepackage{xspace}
\usepackage{hyperref}
\usepackage[linesnumbered,ruled,vlined]{algorithm2e}
\SetKwInput{KwInput}{Input}                
\SetKwInput{KwOutput}{Output}              

\newcommand{\etal}{\textit{et al}.}

\newcommand{\eg}{\textit{e}.\textit{g}.}

\newcommand{\initialmodel}{\textsc{Model-Baseline}\xspace} 
\newcommand{\uvmodel}{\textsc{Model-UV}\xspace}
\newcommand{\deformablemodel}{\textsc{Model-Deformable}\xspace}

\newcommand{\graphmodel}{\textsc{Model-Graph}\xspace}
\newcommand{\gcnmodel}{\textsc{Model-GCN}\xspace}
\newcommand{\nerfmodel}{\textsc{Model-nerf}\xspace}
\newcommand{\gcnresmlp}{\textsc{Model-GGAN}\xspace}
\newcommand{\uvtexture}{\textsc{TextureUV}\xspace}
\newcommand{\vertextexture}{\textsc{TextureVertex}\xspace}
\newcommand{\facetexture}{\textsc{TextureFace}\xspace}

\def\shownotes{1}  \ifnum\shownotes=1
\newcommand{\authnote}[2]{$\ll$\textsf{\footnotesize #1 notes: #2}$\gg$}
\else
 \newcommand{\authnote}[2]{}
\fi

\DeclareUnicodeCharacter{2212}{-}

\usepackage{xcolor}         

\begin{document}
%
\title{Texture Generation Using A Graph Generative
Adversarial Network And Differentiable Rendering}
\titlerunning{Graph Generative Adversarial Network}
%

\author{Dharma KC\inst{1}\orcidID{0000-0003-0676-2391} \and
Clayton T. Morrison\inst{1}\orcidID{0000-0002-3606-0078}\and
Bradley Walls\inst{2}\orcidID{0000-0002-5484-7587}}

%
%

\institute{The University of Arizona, Tucson AZ 85721, USA \email{\{kcdharma,claytonm\}@arizona.edu}
\and
Aret{\'e} Associates, Tucson AZ 85712, USA 
\email{bwalls@arete.com}}

\maketitle              
\begin{abstract}
    Novel photo-realistic texture synthesis is an important task for generating novel scenes, including asset generation for 3D simulations. However, to date, these methods predominantly generate textured objects in 2D space. If we rely on 2D object generation, then we need to make a computationally expensive forward pass each time we change the camera viewpoint or lighting. Recent work that can generate textures in 3D requires 3D component segmentation that is expensive to acquire. In this work, we present a novel conditional generative architecture that we call a {\em graph generative adversarial network (GGAN)} that can generate textures in 3D by learning  object component information in an unsupervised way. In this framework, we do not need an expensive forward pass whenever the camera viewpoint or lighting changes, and we do not need expensive 3D part information for training, yet the model can generalize to unseen 3D meshes and generate appropriate novel 3D textures.
    We compare this approach against state-of-the-art texture generation methods and demonstrate that the GGAN obtains significantly better texture generation quality (according to Fr{\'e}chet inception distance). We release our model source code as open source.\footnote{\url{https://github.com/ml4ai/ggan}}
    
\end{abstract}
\section{Introduction}
Synthesizing novel photorealistic textures for 3D mesh models is an important task for the generation of novel scenes in static images or realistic 3D simulations. Such generated textures can be applied to 3D mesh models and rendered with different lighting conditions and camera angles quite easily. The generative adversarial network (GAN) framework~\cite{goodfellow2014generative} is a promising approach to training models capable of novel image generation. However, extending the GAN framework to support texture generation in 3D that can generalize to novel, previously unseen 3D models poses interesting challenges. Generating textures in 2D space ($u, v$ coordinate system) and then wrapping to 3D mesh models won't generalize to unseen meshes because the $UV$ mapping function is different for different 3D meshes. However, humans are able to identify the components of a 3D object and could texture them consistently.
This raises an interesting research question: can we design an algorithm that can generate realistic textures for unseen 3D mesh models by identifying object components as humans do?
We present here a system that addresses this challenge. Our model can learn to distinguish 3D part information shared across instances of an object class (\eg, wheels, doors, hood, windows, tail, headlights, etc. of a car) in an unsupervised way that supports generating specific texture features for these components. Recent work presented a new model called TM-NET~\cite{gao2021tm} that can generate textures in 3D but requires prior supervised segmentation of 3D parts, while our model can identify 3D part information in an unsupervised way. Another closely related work to ours is~\cite{raj2019learning}, but they work on 2.5D space rather than on the original 3D space, forcing us to make an expensive forward pass each time the viewpoint changes. We use PyTorch~\cite{paszke2019pytorch} and PyTorch3D~\cite{ravi2020accelerating} for the implementation of our system and use the ShapeNet dataset~\cite{chang2015shapenet} to train and evaluate our framework. We adopt the commonly used Fr{\'e}chet Inception Distance (FID)~\cite{heusel2017gans} to assess the quality of textures generated by 
model. FID is typically applied to 2D images. To extend the FID measure to texture map generation for 3D models, we apply the generated texture map that is to be evaluated to the given 3D model and render it from multiple views, producing multiple 2D images, and the FID scores across these images are aggregated to produce a summary FID score. The major contributions of our paper are as follows:
\begin{itemize}
    \item We present a simple solution to the challenging problem of 3D texture generation for 3D mesh models rather than 2D or 2.5D images.
    \item Our framework is capable of unsupervised learning of part information shared across a class of objects, therefore avoiding a costly, separate supervised learning task in order to learn textures appropriate to object parts.
    \item We present a thorough review, analysis, and evaluation of various techniques that can be used for texture generation for 3D meshes and demonstrate the advantages and disadvantages of each.
\end{itemize}

\section{Related work}
In this section, we describe five threads of work related to our problem and proposed framework.
\subsection{Generative adversarial networks}
{\em Generative adversarial networks} (GANs)~\cite{goodfellow2014generative} are known for their ability to generate photorealistic images with very high resolution~\cite{karras2020analyzing}. The GAN framework consists of a generator $\mathcal{G}$ and discriminator $\mathcal{D}$. The generator $\mathcal{G}$ attempts to generate realistic images that can fool the discriminator. At the same time, the discriminator $\mathcal{D}$ tries to predict whether the image is ``real'' (comes from the real data distribution) or ``fake'' (generated by the generator). This constitutes a two-player minimax game with the following value function: 
\begin{equation}
    \begin{split}
        \mathcal{V}(\mathcal{G}, \mathcal{D}) &= \mathbb{E}_{\boldsymbol{x} \sim p_{data}(\boldsymbol{x})} [logD(\boldsymbol{x})] \\
        & + \mathbb{E}_{z\sim p_{z}(\boldsymbol{z})} [log(1 - \mathcal{D}(\mathcal{G}(\boldsymbol{z})))]
    \end{split}
\end{equation}

\noindent Here, $\boldsymbol{x}$ denotes a sample from a real distribution, $p_{data}$, and $\boldsymbol{z}$ denotes a ``noise'' vector from distribution $p_{z}$. $D(\boldsymbol{x})$ denotes the probability that $\boldsymbol{x}$ comes from the 
real distribution. 
Multiple applications have adapted GANs for the task of generating realistic images. Specifically, Radford \etal~\cite{radford2015unsupervised} propose deep convolutional GAN (DCGAN), which uses convolutional neural networks (CNNs) to generate low-resolution photorealistic images. Recently, Karras \etal~\cite{karras2019style, karras2020analyzing, karras2021alias} developed a novel architecture for the generator. Arjovsky \etal~\cite{arjovsky2017wasserstein} propose the Wasserstein GAN (WGAN) to improve the stability of training. Xian \etal ~\cite{xian2018texturegan} propose image synthesis with texture, but this only works on 2D images. Mirza \etal~\cite{mirza2014conditional} propose a conditional GAN framework that can generate samples from a specified class. In this framework, the noise vector is combined with a class label to generate a sample image from that class. This work is closely related to our work as we seek to generate a texture conditioned on an input mesh. This work is a simplified version of our problem as they work on 2D images and the combination of the  noise vector with the class label (a one-hot vector) can be easily achieved with a simple concatenation while a simple concatenation of the noise vector with a 3D mesh model is not possible. This makes our problem challenging and requires a new architecture.

\subsection{Differentiable rendering}
\begin{figure}[H]
    \centering
    \includegraphics[width=0.8\textwidth, height=1.5in]{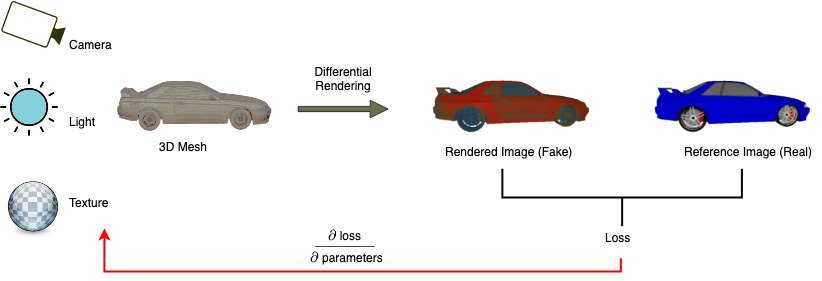}
    \caption{Differentiable rendering}
    \label{fig:dr}
\end{figure}

The second line of work related to ours is {\em differentiable rendering}. Rendering in computer graphics is a process of generating a 2D image from a 3D mesh, light source, camera properties, texture properties, and other scene properties. Classical rendering using rasterization, or ray tracing, is not differentiable. This means we cannot propagate the gradients of the loss in image space (2D space) with respect to mesh properties such as vertices and textures. Given that we want to generate a realistic texture for a given 3D mesh model with supervision from 2D images, we need a way to propagate the gradients of the loss from these projected (rendered) 2D images back to the 3D scene properties. Differentiable rendering is a process that enables backpropagating these gradients from the 2D image loss back into the 3D scene properties. Figure~\ref{fig:dr} illustrates the differentiable rendering part of our architecture. Recent methods propose approximate solutions for making the rendering process differentiable~\cite{loper2014opendr, kato2018neural, liu2019soft, li2018differentiable, ravi2020accelerating}. We use PyTorch3D~\cite{ravi2020accelerating} for differentiable rendering.

\subsection{Texturing}
\label{section:texturing}
{\em Texturing} is the process of applying a texture to a given 3D mesh model. There are multiple ways to apply a texture to a 3D mesh. Given that we want to generate textures for 3D polygonal meshes that can be applied directly to 3D shapes, we have the following options for texturing~\cite{ravi2020accelerating}:
\subsubsection{UV textures:}A {\em UV texture} is a 2D image that can be mapped to 3D mesh model. For UV textures to work, we need a 2D UV-coordinate image and a mapping function that maps every vertex in the 3D object space to a $(u, v)$ coordinate in a 2D UV image. The advantage of this method is that it can represent high-resolution textures, but the mapping function is different for different meshes, making the texturing process hard to generalize across varieties of 3D models. We refer to this method as \uvtexture in the following sections.
\subsubsection{Vertex textures:} A {\em vertex texture} defines a texture per vertex (e.g. r, g, b color). If the mesh has $V$ vertices, and the dimension of texture per vertex is $D$, the texture can be represented by a tensor of shape $(V, D)$ for a given 3D mesh. In this approach, the texture within faces between vertices has to be interpolated from vertex textures. This makes the vertex texture suitable only for low-resolution textures. We refer to this method as \vertextexture in the following sections.
\subsubsection{Face textures:} The {\em face texture} method defines a separate texture per face. The texture per face can be an $R x R$ dimensional texture, where $R$ is the resolution of a texture image of a single face. This can be modeled using a $(F, R, R, D)$ tensor where $F$ is the number of faces, $R$ is the resolution of a texture and $D$ is the dimension of a texture (\eg, $D$ = 3: r, g, b colors). This allows us to learn very high-resolution textures. We refer to it as \facetexture in the following sections.

\subsection{Deformable models}
\label{section:deformable_models}
\begin{figure}[H]
    \centering
    \includegraphics[width=0.6\textwidth, height=1.9in]{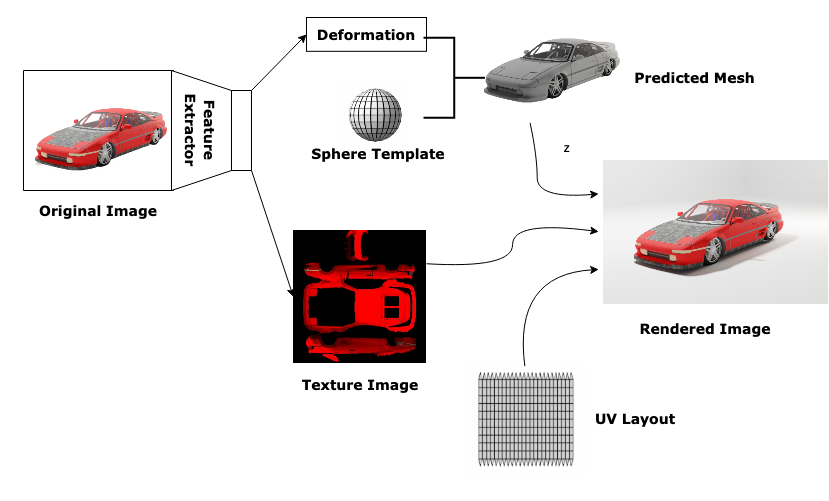}
    \caption{General architecture of deformable models}
    \label{fig:deformable}
\end{figure}

This family of work learns to generate the 3D mesh along with textures from 2D images. The method generally starts with a fixed geometry (e.g. sphere) and a fixed UV mapping. Given input images, the model extracts information from these images, represented as a latent vector. This vector is then used to predict the deformation of the vertices of the sphere template to approximate the 3D mesh and the texture image. Then the estimated 3D shape, estimated texture, and fixed UV map go through a differentiable rendering step to generate a 2D image. The main idea then is to make these generated images similar to the original images, which can be achieved using reconstruction loss and adversarial loss. Figure~\ref{fig:deformable} shows the architecture.
Recent work has explored variations of this idea and has achieved good results~\cite{pavllo2021learning, pavllo2020convolutional, goel2020shape, chen2019learning, henderson2020leveraging, pavllo2020convolutional, zhang2020image, kanazawa2018learning}. This is a really good approach when we don't have ground truth 3D meshes. But, when the 3D meshes are available, as in our case, the major disadvantage of this method is that the predicted mesh tends to be relatively poor quality compared to the ground-truth mesh.

\subsection{Graph neural networks}
Graph neural networks (GNNs) are powerful models for learning from graph-structured data. They work on the theory of message passing, where a node gets some information from its neighbors and updates its state. 
Consider a graph $\mathcal{G} = (\mathcal{V}, \mathcal{E})$, where $\mathcal{V}$ is the set of nodes and $\mathcal{E}$ is the set of edges. Let, $\mathcal{X} \in \mathbb{R}^{|v| * d}$ be the set of node features where each node $v \in \mathcal{V}$ has a $d$ dimensional feature. The $k^{th}$ message passing iteration of a GNN can be modeled as a variation of the following equation~\cite{grlbook}:
\begin{equation}
\begin{split}
        h_v^{(k+1)} &= \texttt{update}^{(k)}(h_v^{(k)}, \texttt{aggregate}^{(k)}(h_u^{(k)}),\\\
        &= \texttt{update}^{(k)}(h_v^{(k)}, \boldsymbol{m}_{\mathcal{N}(v)}^{(k)}) \quad \forall u \in \mathcal{N}(v))
\end{split}
\end{equation}
Here, $\mathcal{N}(v)$ denotes the neighbors of node $v$. At any iteration of the GNN, the \texttt{aggregate} function takes the embedding of the neighbors of node $v$ and combines them into one embedding vector. The \texttt{update} function takes the embedding of the node $v$ at the previous time step and the output embedding vector of the  \texttt{aggregate} function to give us the new embedding for the node $v$. Here, \texttt{update} and \texttt{aggregate} can be any differentiable functions. 
In our work here, we convert an input 3D mesh model to a graph and use the power of GNNs to learn latent part information of a class of objects.

\section{Models} 
In this section, we describe multiple different approaches to addressing the problem of how to generate novel but realistic textures for variant 3D meshes of an object class. We discuss the advantages and disadvantages of these methods. In summary:
\begin{itemize}
    \item \initialmodel:
    This model utilizes simple UV mapping.
    We found that this architecture doesn't generalize to unseen 3D meshes.
    \item \uvmodel: 
    This model takes UV layout as extra input information but suffers similar problems to \initialmodel.
    \item \deformablemodel: 
    This method is based on the idea of deformable models. The disadvantage of this method is that the approximated mesh is of low quality compared to the original 3D mesh.
    \item \graphmodel: This model consists of two variants, \gcnmodel, and \gcnresmlp. These models transform an input 3D mesh into a graph and generate a texture conditioned on the graph. The final variant of this model, called \gcnresmlp, produced higher quality and more diverse results than the other previous methods.
\end{itemize}
In the following section, we describe each model in detail.
\subsection{\initialmodel}
\begin{figure}[H]
    \centering
    \includegraphics[width=\textwidth, height=3.2in]{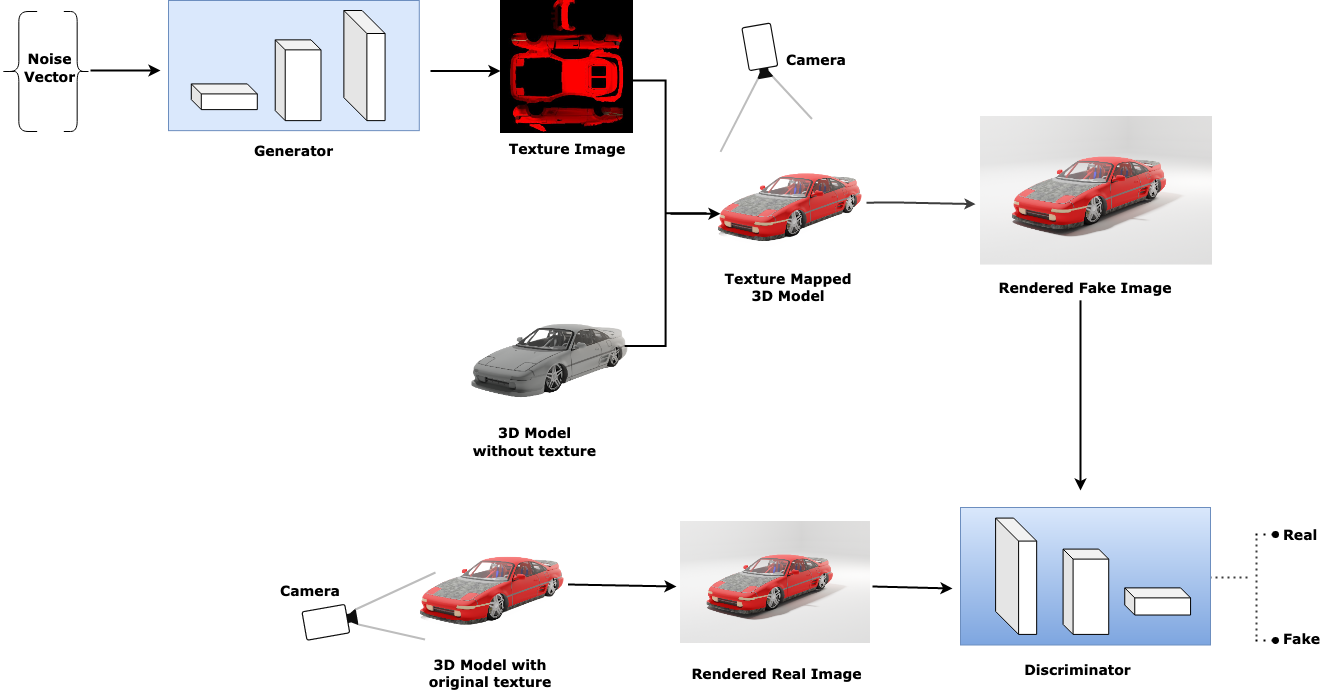}
    \caption{Initial architecture for texture synthesis}
    \label{fig:system}
\end{figure}

In this first model, 
the texturing is performed using the \uvtexture mechanism.
The framework is summarized in Figure~\ref{fig:system}.
For this model, we adapted a deep convolutional generative adversarial network (DCGAN) architecture~\cite{radford2015unsupervised} for the generator and the discriminator. We modified the DCGAN architecture to support the generation of higher-resolution textures. For applying the texture map to these diverse 3D models, we need a way to map the 3D vertices in these models into a 2D texture map (a procedure called UV mapping).  We use the {\em smart UV project} feature of the Blender Python API to automatically generate these mappings for a given 3D model. Recall that our challenge is to adapt the GAN framework so that we can take advantage of training on multiple real-world examples and have the learned generation capability transfer to new 3D objects. However, the UV image for each 3D model has a different coordinate system. This means that, for example, the features of the given 3D model, such as the tires or windshields of different cars, project to different regions in the UV space for each 3D model. 

To establish a baseline, we directly applied the UV mapping
and found that this led the generator to converge to a mean texture across examples, rather than learning how to produce varied textures constrained by the input 3D mesh.
This happens because the generator has no information about the UV mapping function and 3D mesh model to which the texture will be applied.

\subsection{\uvmodel}
To address the above issue, we next explored the idea of injecting the UV map layout into the generator with the hypothesis that the generator might adapt during training in order to learn where the different parts of the given 3D model project in the 2D texture image. 
We used the same texturing mechanism, \uvtexture, as described in the above \initialmodel. The architecture produced results similar to \initialmodel. The reason was the model couldn't learn complicated UV mapping information just from the UV layout image.

\subsection{\deformablemodel}
In this model, we use the idea of deformable models similar to Figure~\ref{fig:deformable}, but with some modifications. We use the same \uvtexture method described above. The main problem with the above two models is that the UV mapping function is different for different 3D mesh models, making it harder for the generator to learn features that can work across different 3D mesh models. To mitigate this problem, we explored the idea of starting from a common mesh model with a fixed UV mapping. We used a sphere template 3D model as the starting point and used azimuth and elevation as the UV map function. We then used 3D chamfer loss~\cite{fan2017point} to predict the deformations of the sphere template to approximate the 3D mesh. This model more directly addresses our overall challenge by learning a generalized mapping from the space of textures to different 3D meshes enabling us to swap the texture learned from one model to another. However, the generated textures must still be applied to the approximate model, which reduces the quality of the 3D mesh model and the texture. The distortions in the model shape mesh are significant as demonstrated in the example in Figure~\ref{fig:swap_2}. Another disadvantage of this method is that we need to approximate the deformation for every new 3D model, creating extra computational overhead for training and inference.

\subsection{\graphmodel}
\begin{figure}[H]
    \centering
    \includegraphics[width=\textwidth, height=3.7in]{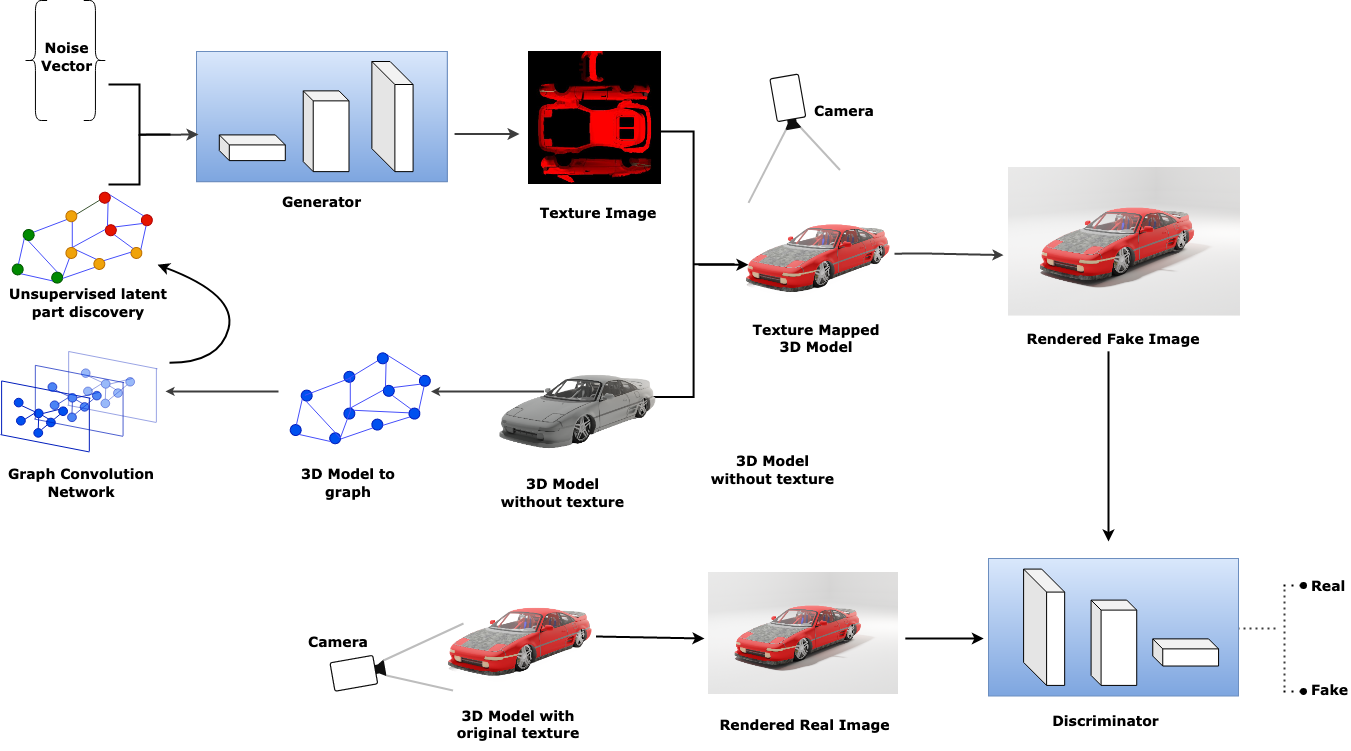}
    \caption{Graph-based methods for texture synthesis}
    \label{fig:system_gnn}
\end{figure}

In this section, we describe our architecture that incorporates information from the 3D mesh model to guide the generator. We first convert the input 3D mesh model into a graph by taking each face as a node in the graph and connecting neighboring faces using graph edges. The graph neural network is then used to learn a latent representation of the structural components of the given 3D model. In the latent representation, the topological features of the 3D mesh graph can be clustered, so as to learn features that could correspond to structural components that tend to share texture properties, such as wheels, windows, lights, and hood of a car. In turn, this latent component representation can then provide an inductive bias for the generator to produce a texture for the given 3D mesh model. The architecture is shown in Figure~\ref{fig:system_gnn}. 
An interesting aspect of this design is that the generator can take the unsupervised latent part representation as node features and combine it with the input noise vector to generate a texture for the particular 3D mesh. Node features is a 2D tensor of shape $v, f$ where $v$ is the number of nodes and $f$ is the dimension of the node feature.
The noise is a 1D vector of shape $d$. We sample a noise vector $z \in \mathbb{R}^d$ from a multivariate normal distribution. This $d$ dimensional noise vector is then replicated to have a shape of $v * d$. This allows our model to process 3D mesh models with a different number of nodes. This noise tensor is then concatenated with the node feature tensor $v * f$. The concatenated tensor is then input to the generator (e.g. MLP) that, in turn, generates the textures for the given 3D mesh model. We use \facetexture for texturing as it enables us to generate higher quality textures than \vertextexture.
We represent the faces of the 3D mesh as the nodes in the graph. The $x, y, z$ face position and its normal ($n_x, n_y, n_z$) form the initial node features. We use a graph convolutional neural network (GCN)~\cite{kipf2016semi} to learn the latent part representation as shown in Figure~\ref{fig:system_gnn}. The generator generates a tensor of shape $F * 3$, where $F$ is the number of faces (nodes in the graph), and 3 represents the three $r, g, b$ colors (texture) per face. We use \facetexture for texturing the 3D mesh. We explored the following two variants of this \graphmodel that differ only in the design of the generator. \gcnmodel uses GCN~\cite{kipf2016semi} as a generator to generate the texture from a combination of latent part representation and noise vector. And \gcnresmlp uses a multi-layer perceptron with residual connections~\cite{he2016deep} as a generator. \gcnresmlp is our best-performing model.

\section{Experiments}
\label{section:experiments}
We use the ShapeNet~\cite{chang2015shapenet} car data set for all of our experiments. This data set consists of a total of 3,514 3D mesh models of cars with textures. We use 3314 mesh models for training, 100 mesh models for validation, and 100 mesh models for testing. The features extracted from intermediate layers of the pre-trained deep neural networks are known to correspond to the perceptual metrics of human vision~\cite{johnson2016perceptual, zhang2018unreasonable}. We found that incorporating this perceptual loss into the generative adversarial loss improved the qualitative appearance of the generated textures.
Thus our overall loss function is as follows:
\begin{equation}
    \texttt{Loss(L)}  = \texttt{gan\_loss}  + \lambda * \texttt{perceptual\_loss}
\end{equation}

We use the validation dataset to select the best value of $\lambda$. We use the library of Zhang \etal~\cite{zhang2018unreasonable} to extract features from the intermediate layer of pre-trained AlexNet architecture~\cite{krizhevsky2012imagenet} that are in turn used to calculate the perceptual loss. All of the models are trained with the above loss function. At each minibatch iteration, we render a 3D mesh with real and synthetic textures from eight different viewpoints. The loss is calculated from these real and synthetic (``fake'') images. We use a learning rate of 0.0001 for both the generator and the discriminator. We use a hidden size of 64 for both GNN and the MLP generator. Here, hidden size is the dimension that's being used to project the node features of the graph. We render images of size $512 \times 512$ from the differentiable renderer. We use the Adam optimizer~\cite{kingma2014adam} for training all of our models. We used $d = 16$ for the random noise vector. For \graphmodel variants, we use 3 graph convolution layers~\cite{kipf2016semi} for learning the latent part representation. Each convolutional layer has a hidden size of 64. The noise vector is concatenated to the output of the last graph convolutional layer. We use \facetexture to texture the 3D mesh model. For the \gcnmodel, we use a 7-layered GCN~\cite{kipf2016semi} as a generator with a hidden size of 64. The generator does not use residual connections~\cite{he2016deep}. For the \gcnresmlp architecture, we use a 7-layered MLP as a generator with a hidden size of 64. The generator uses residual connections~\cite{he2016deep}.

\section{Results}
\begin{figure}[ht]
\centering
\begin{minipage}{.5\textwidth}
  \centering
  \includegraphics[width=.5\linewidth]{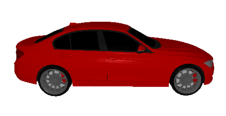}
  \captionof{figure}{Original mesh with original\\ texture}
  \label{fig:swap_1}
\end{minipage}%
\begin{minipage}{.5\textwidth}
  \centering
  \includegraphics[width=.5\linewidth]{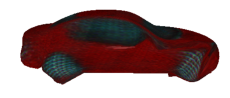}
  \captionof{figure}{approximate mesh with learned texture using \deformablemodel}
  \label{fig:swap_2}
\end{minipage}
\end{figure}

\begin{figure}[h]
  \subfloat[]{%
  \begin{minipage}{\linewidth}
  \includegraphics[width=.3\linewidth]{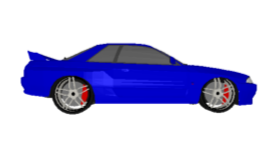}\hfill
  \includegraphics[width=.3\linewidth]{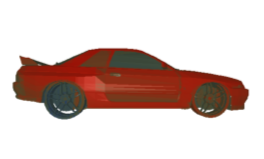}\hfill
  \includegraphics[width=.3\linewidth]{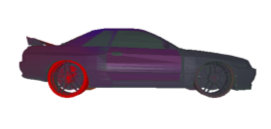}\hfill
  \end{minipage}%
  }\par
  \subfloat[]{%
  \begin{minipage}{\linewidth}
  \includegraphics[width=.3\linewidth]{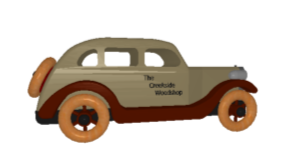}\hfill
  \includegraphics[width=.3\linewidth]{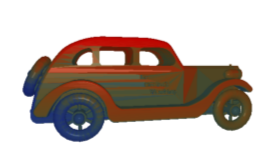}\hfill
  \includegraphics[width=.3\linewidth]{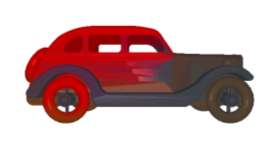}\hfill
  \end{minipage}%
  }\par
  \subfloat[]{%
  \begin{minipage}{\linewidth}
  \includegraphics[width=.3\linewidth]{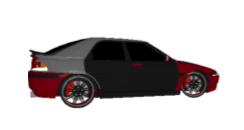}\hfill
  \includegraphics[width=.3\linewidth]{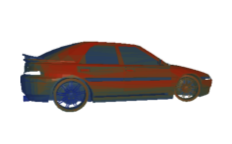}\hfill
  \includegraphics[width=.3\linewidth]{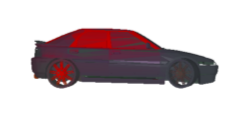}\hfill
  \end{minipage}%
  }\par
  \subfloat[]{%
  \begin{minipage}{\linewidth}
  \includegraphics[width=.3\linewidth]{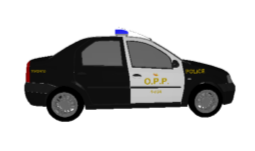}\hfill
  \includegraphics[width=.3\linewidth]{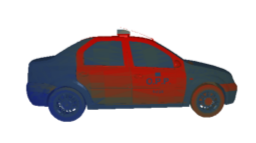}\hfill
  \includegraphics[width=.3\linewidth]{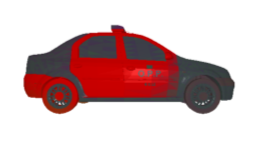}\hfill
  \end{minipage}%
  }
  \caption{First column: original images, second column: \gcnresmlp, third column: \gcnmodel}
  \label{fig:results}
\end{figure}

The \initialmodel and \uvmodel were only able to learn to generate textures for a single mesh model and were not able to generalize across unseen 3D mesh models. The \deformablemodel was able to generate a texture for unseen 3D mesh models, but the quality of approximated mesh and texture was not good, as demonstrated in contrast between
Figures~\ref{fig:swap_1} and \ref{fig:swap_2}. Moreover, the approximation of the 3D mesh model created extra computational overhead. Thus we didn't move forward with this approach for the full ShapeNet car experiment.
\vspace{-2em}
\vspace{-1em}
The graph-based models based on the \graphmodel architecture were able to learn textures across different 3D mesh models. 
This general approach has multiple advantages compared to existing solutions that generate textures in 2D. First, the model is able to learn about the parts of the given 3D mesh model in an unsupervised way. This removes the effort and cost required for manual labeling of the 3D part segmentation. Second, the approach generates textures in 3D, so that a texture can be applied once and the 3D model can be viewed from multiple directions and under multiple light conditions without a need to generate texture each time we change these parameters.
We evaluated these models by applying the synthetic texture generated from respective models and rendering them from multiple viewpoints. We then calculated the FID score based on these projected images and actual original images rendered from the same views with the original texture. 
The \initialmodel, \uvmodel, and \deformablemodel architectures were not suitable for learning textures across different 3D mesh models, so we did not compute FID scores for these. Table~\ref{table:results-table} shows the average FID values (lower is better) for different models per 3D mesh model.
For further comparison, we also experimented with a simple variant of the NeRF~\cite{mildenhall2020nerf} model as a generator (\nerfmodel), but it did not produce results as good as the graph neural network approaches. The low quality of results is reasonable because it doesn't have a way for learning part information like our \graphmodel.
\vspace{-1em}
\begin{table}[H]
\caption{FID scores on the test dataset}
\centering
\begin{tabular}{ ||c|c|| } 
 \hline
 \emph{Model} & \emph{FID}  \\
 \hline
 \hline
 \gcnmodel & 0.75 \\
 \hline
 \nerfmodel &  0.93\\ 
 \hline
 \textbf{\gcnresmlp} & $\boldsymbol{0.70}$ \\
 \hline
\end{tabular}
\label{table:results-table}
\end{table}
Figure~\ref{fig:results} shows a set of selected examples of rendered images generated from different models with a fixed viewpoint. Textures are applied to the 3D mesh models and rendered as projected 2D images for visualization. The first column shows the images rendered with original textures, the second column shows the images rendered with textures generated from the \gcnresmlp model, and the third column shows the images rendered with textures generated from the \gcnmodel. In Figure~\ref{fig:results}, we observe that the \gcnmodel lacks diversity in the generated images: it produces images with the same texture for every random noise input. We hypothesize that this is due to the over-smoothing problem observed in GNNs~\cite{cai2020note, chen2020measuring} as the model uses GNN-only layers for the generator. Finally, the model \gcnresmlp (GGAN) produces images (second column, Figure~\ref{fig:results}) that respect the object boundaries, are visually better than other models and are diverse (the model produces new textures on each run with different random noise input). Some of the images generated from our final model \gcnresmlp (third row, second column) look even better than the original image itself (third row, first column).

\section{Conclusion}
In this work, we have presented and evaluated the graph generative adversarial network (GGAN), a new architecture that can learn to generate a texture for a given 3D mesh with high fidelity and that can learn 3D part information in an unsupervised way. We think GGAN will be useful in various domains to generate graph-structured representation.
However, there are multiple directions for improvement. The first important research direction for future work is to introduce symmetry constraints on the system such that all components with symmetrical structures will generate textures that respect symmetries. Second, we want to increase the diversity of the generated textures. Third, we want to improve the controlled synthesis of part-specific textures. Another important research direction would be to incorporate encoder-decoder graph architectures~\cite{gao2019graph} within our framework. Another important direction would be to couple with a semi-supervised labeling approach. Finally, we would like to explore the use of flow-based models~\cite{weng2018flow, rezende2015variational} and diffusion models~\cite{weng2021diffusion, dhariwal2021diffusion} for the generation of texture within our current framework.

\bibliographystyle{splncs04}
\bibliography{mybib}
\end{document}